\documentclass[10pt,twocolumn,letterpaper]{article}

\usepackage{iccv}              

\definecolor{iccvblue}{rgb}{0.21,0.49,0.74}
\usepackage[pagebackref,breaklinks,colorlinks,allcolors=iccvblue]{hyperref}
\usepackage{algorithm2e}

\usepackage[accsupp]{axessibility}  

\def\paperID{10} 
\def\confName{ICCV}
\def\confYear{2025}

\title{Probabilistic Domain Adaptation for Biomedical Image Segmentation}

\author{Anwai Archit{$^{1,2}$} and Constantin Pape{$^{1,2,3,4}$}\\
{\small {$^{1}$} Georg-August-University G\"ottingen, Institute of Computer Science}\\
{\small {$^{2}$} Campus Institute Data Science (CIDAS), G\"ottingen, Germany}\\
{\small {$^{3}$} CAIMed - Lower Saxony Center for AI \& Causal Methods in Medicine, G\"ottingen}\\
{\small {$^{4}$} Cluster of Excellence Multiscale Bioimaging (MBExC), Georg-August-University G\"ottingen}\\
}

\begin{document}
\maketitle

\begin{abstract}
Segmentation is a crucial analysis task in biomedical imaging. Given the diverse experimental settings in this field, the lack of generalization limits the use of deep learning in practice. Domain adaptation is a promising remedy: it involves training a model for a given task on a source dataset with labels and adapts it to a target dataset without additional labels. We introduce a probabilistic domain adaptation method, building on self-training approaches and the Probabilistic UNet. We use the latter to sample multiple segmentation hypotheses to implement better pseudo-label filtering. We further study joint and separate source-target training strategies and evaluate our method on three challenging domain adaptation tasks for biomedical segmentation.
\end{abstract}

\section{Introduction}
\label{sec:intro}

Deep learning has emerged as the leading approach for various image analysis tasks, including segmentation in biomedical imaging. However, its practical application is often hindered by limited generalization and the need for large labeled training datasets, typical in supervised learning. These drawbacks are particularly challenging in biomedical imaging, where many different experimental setups and imaging modalities exist. Domain adaptation offers a promising solution by adapting a model trained on a labeled dataset (”source”) for a specific task, for example cell segmentation, to perform the same task effectively on a new dataset (”target”) from a different domain. We present a novel domain adaptation method that integrates self-training techniques with probabilistic segmentation and demonstrate its effectiveness for semantic segmentation in biomedical images. Our proposed method shows significant promise in improving generalization across domains and reducing the labeling effort in practice.

Self-training for domain adaptation builds on ideas from semi-supervised learning. These approaches train a model jointly on the labeled and unlabeled data. On the unlabeled data, predictions from a pretrained version of the model (the ”teacher”) are used as so-called pseudo-labels for the model (the ”student”). A popular self-training method is \textit{Mean Teacher} \cite{mean_teacher}, which uses the exponential moving average (EMA) of the student weights to update the teacher weights. Recently \textit{FixMatch} \cite{fixmatch} has gained popularity. Its teacher and student share weights, but weak and strong augmentations prevent collapse of the model predictions. Both approaches have been studied for semi-supervised classification as well as segmentation, for example in \textit{UniMatch} \cite{unimatch}. Furthermore, self-training can be extended to domain adaptation, see for example \textit{AdaMatch} \cite{adamatch}.

\begin{figure*}[htbp]
  \centering
  \includegraphics[width=\linewidth]{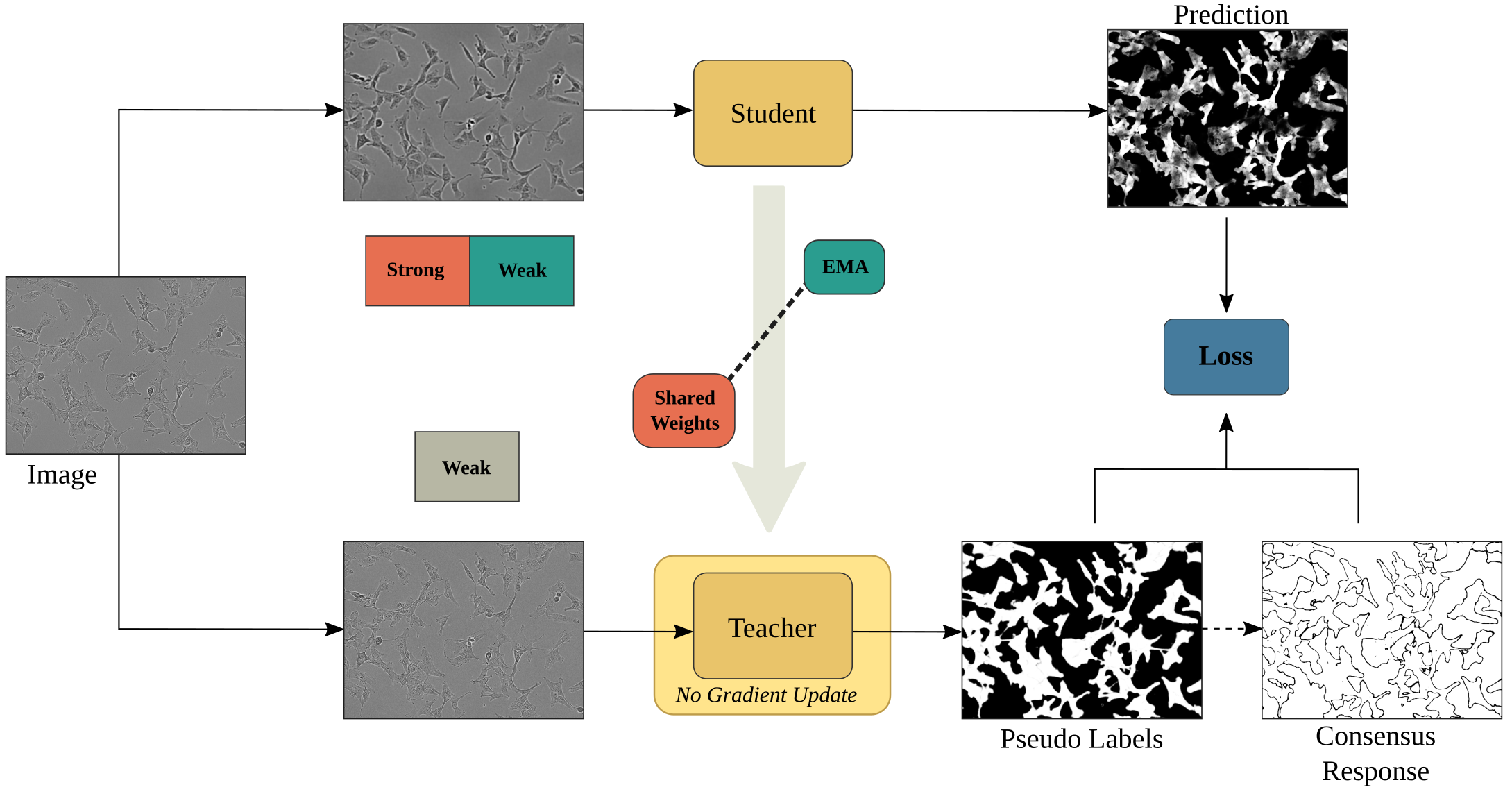}
  \caption{Self‐training and probabilistic segmentation: the teacher predicts
  pseudo-labels, which are compared with the student predictions. Multiple
  samples from the probabilistic teacher filter the pseudo-labels. Student and
  teacher either share weights (\textit{FixMatch, \textcolor{red}{red}}) or the teacher weights
  are the EMA of the student weights (\textit{MeanTeacher, \textcolor{teal}{green}}). Their inputs
  receive different augmentations: weak (\textit{MeanTeacher}) and weak + strong
  (\textit{FixMatch}).}
  \label{fig:domain_adaptation}
\end{figure*}

We study three key design choices for applying self-training to domain adaptation: (i) how to generate the pseudo-labels, (ii) how to filter them and (iii) how to orchestrate source and target training. \textit{Mean Teacher}, \textit{FixMatch} and others have mostly focused on (i), by studying different student-teacher set-ups and different augmentation strategies. For (ii) most approaches either do not filter the pseudo-labels (e.g. \textit{MeanTeacher}), or filter them based on confidences derived from deterministic model predictions (e.g. \textit{FixMatch}). However, regular deep learning methods are known to be badly calibrated and hence their predictions do not yield reliable confidence estimates. Instead, we propose to use the \textit{Probabilistic UNet} (PUNet) \cite{punet}, a probabilistic segmentation method, to derive better confidence estimates, and use them to filter pseudo-labels. Fig. \ref{fig:domain_adaptation} shows an overview of our method. We also investigate two different strategies for (iii): separate source training and subsequent adaptation (two stages) versus joint training on source and target (single stage). We evaluate our method on three challenging domain adaptation tasks in biomedical segmentation: cell segmentation in livecell microscopy, mitochondria segmentation in electron microscopy (EM) and lung segmentation in X-Ray. Our method compares favorably to strong baselines.
Our code is available at \url{https://github.com/computational-cell-analytics/Probabilistic-Domain-Adaptation}.

\section{Related Work}

\textit{Self-training}: These methods train a model on unlabeled data (either from the same data domain in semi-supervised learning or from a different data domain in domain adaptation) on a version of its own predictions, employing different strategies to prevent collapse (e.g. all predictions are zeros). They are among the most popular approaches in semi-supervised learning; examples include \textit{Mean Teacher} \cite{mean_teacher}, \textit{ReMixMatch} \cite{remixmatch} and \textit{FixMatch} \cite{fixmatch}. They are often studied for classification tasks, but have also been successfully applied to semantic segmentation, see for example \textit{PseudoSeg} \cite{pseudoseg} and \textit{UniMatch} \cite{unimatch}. Similar methods have emerged for domain adaptation: a unified framework for semi-supervision and domain adaptation has been proposed by \textit{AdaMatch} \cite{adamatch}. Self-training methods for domain adaptation in segmentation have for example been introduced by Zou et al. \cite{zou_domain}, which uses class-balanced pseudo-labeling, and Mei et al. \cite{mei_domain}, which uses instance adaptive pseudo-labeling. Applications to segmentation in biomedical images include Shallow2Deep \cite{shallow2deep}, which uses intermediate predictions from a shallow learner for pseudo-labels and SPOCO \cite{spoco}, which uses a per-instance loss and self-training to learn from sparse instance segmentation labels.
Recently, a tool for the segmentation of synaptic structures, SynapseNet \cite{synapsenet}, has also adopted this approach to improve generalization of its models.

\textit{Probabilistic segmentation}: Probabilistic segmentation methods learn a distribution over segmentation results and enable sampling from it. Early approaches have implemented this through test-time dropout \cite{test-time-dropout}, recent approaches such as the PUNet \cite{punet} and its hierarchical extension \cite{hpunet, phiseg}, use a conditional variational autoencoder. With the emerging success of diffusion probabilistic models, one could use approaches such as MedSegDiff \cite{medsegdiff} for sampling multiple hypotheses segmentation. Prior work has introduced strategies for filtering pseudo-labels using probabilistic segmentation for semi-supervised learning; either using the dropout approach \cite{uncertainty-dropout1, uncertainty-dropout2} or inconsistencies between different models \cite{inconsistency}. 

\begin{figure*}[htbp]
  \centering
  \includegraphics[width=\linewidth]{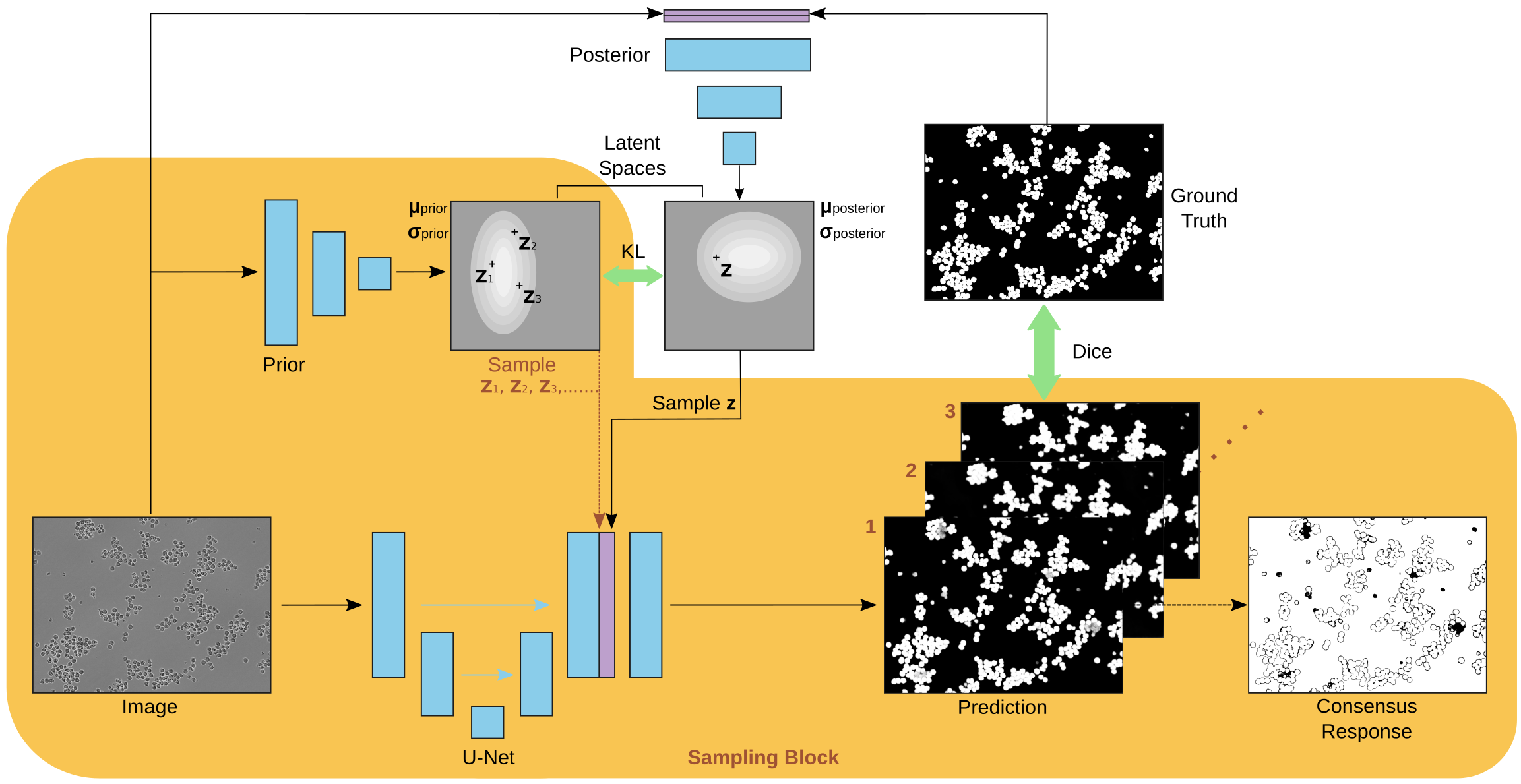}
  \caption{PUNet setup: during training, both prior and posterior encoders predict parameters for sampling latents. Features from the UNet are concatenated with the posterior latents for prediction. The KL divergence minimizes the distance between prior and posterior predictions, while the Dice loss compares ground-truth segmentation and predicted labels. In inference (highlighted in \textcolor{orange}{orange}), the prior encoder is used for sampling.}
  \label{fig:punet}
\end{figure*}

\section{Methods}

We propose a new approach to domain adaptation for semantic segmentation. We first introduce the self-training methodology, review its application to domain adaptation, review probabilistic segmentation, and explain how we combine these elements for probabilistic domain adaptation. Exact definitions for all terms are given in App. \ref{app_definitions}.

\textit{Self-training for semi-supervised learning}: Self-training approaches train a model using the labeled samples to compute the supervised loss term $L_s$ and the unlabeled samples to compute the unsupervised loss term $L_u$. The term $L_u$ is formulated based on a student-teacher set-up, which uses the teacher predictions as pseudo-labels. For the student the same model that is trained via $L_s$ is used, the teacher is derived from it (e.g. by weight sharing or EMA of the weights). To avoid over-fitting to the teacher predictions, the inputs to the student and teacher are transformed with different augmentations and the pseudo-labels are filtered and/or post-processed.
Denoting the supervised inputs and labels as $x_s$ and $y_s$, the inputs without labels as $x_u$, the student and teacher model as $s$ and $t$, the augmentations for the student and teacher as $\tau_s$ and $\tau_t$, the function that filters the pseudo-labels as $f$, and the function that post-processes them as $p$, we can write the generation of pseudo-labels $\hat{y}_u$ and the combined loss $L$ as:

\begin{equation} \label{eq_pseudolabels}
    \hat{y}_u = p(t(\tau_t(x_u))),
\end{equation}

\begin{equation} \label{eq_loss}
    L = L_s(s(x_s), \, y_s) + L_u(f(s(\tau_s(x_u)), \, \hat{y}_u)).
\end{equation}

The operations in Eq.~\ref{eq_pseudolabels} are not taken into account for the gradient computation. We obtain \textit{MeanTeacher} from Eq.~\ref{eq_pseudolabels} and \ref{eq_loss} by choosing $\tau_s$ and $\tau_t$ to be weak augmentations from the same distribution, choosing both $f$ and $p$ to be the identity function (i.e. pseudo-labels are neither filtered nor post-processed), and computing the teacher weights $w_t$ as the EMA of the student weights $w_s$: $w_t = \alpha \, w_t + (1 - \alpha) \, w_s$ with $\alpha$ close to 1. We obtain \textit{FixMatch} \cite{fixmatch} by choosing $\tau_s$ to be strong augmentations, $\tau_t$ to be weak augmentations, $t$ to share weights with $s$, $f$ to filter out predictions with a probability below a given threshold, and $p$ to set the predicted values to 1 if they are above that threshold. We apply \textit{MeanTeacher} and \textit{FixMatch} to semantic segmentation. Our modifications for this task, including the choices for $L_s$ and $L_u$, are given later.

\paragraph{Self-training for domain adaptation:} There are two possible strategies for applying the self-training procedures from the previous paragraph to domain adaptation: in the \textit{joint strategy} the model is trained according to Eq.~\ref{eq_pseudolabels} and \ref{eq_loss} using the source data as labeled samples $(x_s, y_s)$ and the target data as unlabeled samples $x_u$. In the \textit{separate strategy} the model is pre-trained on the source domain, using the supervised loss $L_s$, and then adapted to the target domain, using the unsupervised loss $L_u$.
The \textit{joint strategy} is a direct extension of semi-supervised learning to domain adaptation, as \textit{AdaMatch} \cite{adamatch} has shown through a unified formulation of both learning paradigms.
In contrast, the \textit{separate strategy} requires two training stages. It has the practical advantage that only a single model has to be pre-trained per task, which can then be fine-tuned for a new target dataset without access to the source data and with a small computational budget. In comparison, the \textit{joint strategy} requires training a model for each source-target pair, with access to the source data. Here, we compare both domain adaptation strategies.

\paragraph{Probabilistic segmentation:} Probabilistic segmentation methods have been introduced to account for the inherent uncertainty of the segmentation tasks: in many cases there is not a single "true" segmentation solution, but rather a distribution. For example, in the case of radiology, lesions are often annotated differently by experts, without a definite consensus solution. To model this label uncertainty, probabilistic segmentation methods enable sampling diverse segmentation results instead of providing a deterministic result.
Here, we use the PUNet \cite{punet}: it combines a UNet \cite{unet}, which learns deterministic features, with a conditional variational autoencoder \cite{cvae} that enables sampling. During training it uses a prior encoder, which sees only the images, and a posterior encoder, which sees the images and labels. Both encoders map their respective inputs to the parameters of a multi-dimensional gaussian distribution; sampling from it yields representations in a latent space. The latent samples from the posterior and the UNet features are concatenated and passed to additional layers to obtain the segmentation result. The architecture is trained using a variational loss term $L_{var}$ that minimizes the distance between the distributions of the prior and posterior encoder, and a reconstruction loss term $L_{rec}$ that minimizes the difference between prediction and labels. For inference the prior encoder is used for sampling. See Fig.~\ref{fig:punet} for an overview of the PUNet architecture and training.

Here, we use the PUNet to sample diverse predictions and to implement pseudo-label filtering based on these predictions.
We introduce the \textit{consensus response} $C$. Given predictions $m(x_i)^k_j$ from a PUNet $m$ for pixel $i$, class $k$ and sample $j$, the per pixel response $c_i$ is:
\begin{equation} \label{eq_consensus}  
    c_i = \frac{1}{N} \sum_j^N \, \{1 \, \textrm{if} \, t(x_i)^k_j \ge \theta \, \textrm{for any} \, k \in K\ \, \textrm{else} \, 0\}.
\end{equation}

Here, $\theta$ denotes the threshold parameter, $N$ the number of samples and $K$ the number of classes. This expression generalizes the thresholding approach for a single prediction, which is recovered for $N=1$. See the next paragraph for how $C$ is used for pseudo-label filtering.

\paragraph{Probabilistic domain adaptation:} We combine the elements from the previous paragraphs into a family of methods for probabilistic domain adaptation. We follow the PUNet implementation of \cite{punet}, but use the dice score for $L_{rec}$ instead of the cross entropy. We use the same loss function ($L = L_{var} + L_{rec}$) for both $L_s$ and $L_u$ (see Eq.~\ref{eq_loss}). We do not use pseudo-label post-processing, i.e. $p$ is the identity (cf. Eq~\ref{eq_pseudolabels}).

We study two different settings for self-training: for \textit{MeanTeacher} $\tau_s$ and $\tau_t$ are both weak augmentations and the teacher weights are the EMA of the student weights. For \textit{FixMatch} $\tau_s$ are strong augmentations, $\tau_t$ are weak augmentations, and the student and teacher share weights. We use the consensus response $C$ (Eq.~\ref{eq_consensus}) for filtering the reconstruction loss $L_{rec}$. Here, we explore three different settings: \textit{consensus masking}, which masks the pixels that do not fulfill $c_i = 1$, \textit{consensus weighting}, which weights the loss with the value of $c_i$, and no filtering. The set-up for unsupervised training is shown in Fig. \ref{fig:domain_adaptation}. We also investigate both the \textit{joint} and \textit{separate} domain adaptation strategies. Tab.~\ref{tab_methods} gives an overview of the proposed methods and the corresponding abbreviations. Pseudo-code for the methods can be found in App.~\ref{app_pseudocode}.

\begin{table}[h]
\centering
\small
\caption{Proposed probabilistic domain adaptation methods and abbreviations. In the "Filtering" column, "Cons." is short for consensus, and \textit{None} indicates no pseudo-label filtering used for the particular setting.}
\label{tab_methods}
\begin{tabular}{llcc}
\toprule
Setting & Filtering & FixMatch & MeanTeacher \\
\midrule
Joint & Cons. Masking & $FM^m_j$ & $MT^m_j$ \\
Joint & Cons. Weighting & $FM^w_j$ & $MT^w_j$ \\
Joint & \textit{None} & $FM_j$   & $MT_j$   \\
Separate & Cons. Masking & $FM^m_s$ & $MT^m_s$ \\
Separate & Cons. Weighting & $FM^w_s$ & $MT^w_s$ \\
Separate & \textit{None} & $FM_j$   & $MT_s$   \\
\bottomrule
\end{tabular}
\end{table}

\begin{figure*}[htbp]
  \centering
  \includegraphics[width=\linewidth]{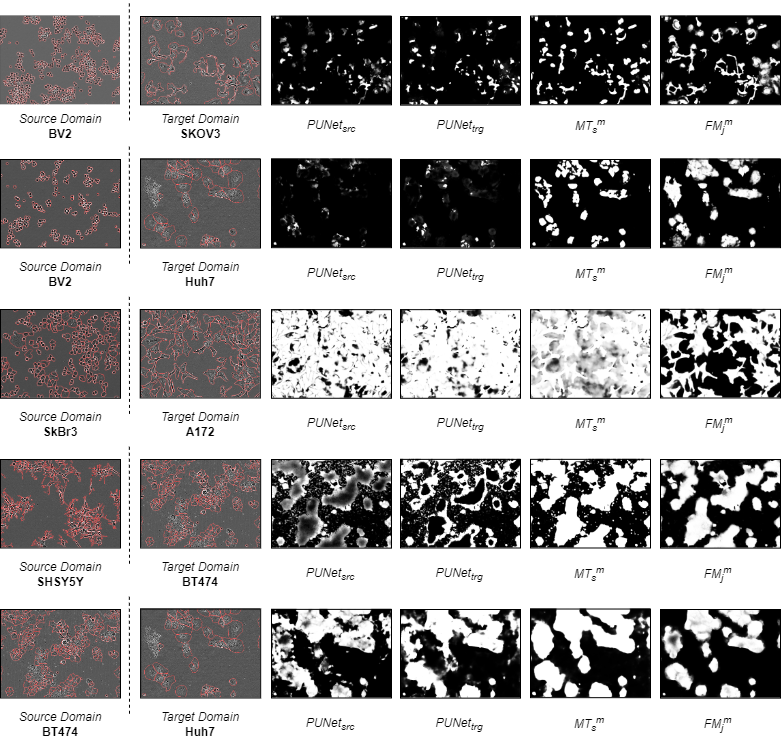}
  \caption{Qualitative LIVECell segmentation results. The left panel shows the labeled source‐domain image; the right panel shows (i) the target‐domain image, with predictions from four methods: (ii) PUNet trained only on the source domain, (iii) PUNet adapted to the target domain, and (iv–v) two self-training approaches.}
  \label{fig:da-livecell}
\end{figure*}

\section{Results}

We perform experiments for three different domain adaptation settings: cell segmentation in livecell microscopy, mitochondria segmentation in EM and lung segmentation in X-Ray. Model implementation and hyper-parameters are documented in detail in App.~\ref{app_method_details}, the datasets in App.~\ref{app_dataset_details}.

\subsection{Domain Adaptation Results}

We perform comprehensive experiments for cell segmentation in livecell microscopy. We use data from LIVECell \cite{livecell}, which contains 5,000 images with cell instance labels for 8 different cell lines. We transform the instance labels to binary masks for semantic segmentation, treat each cell line as a different domain and evaluate the segmentation quality for all source target pairs. In Tab.~\ref{tab1} the averaged dice scores for each source domain applied to the 7 corresponding target domains are given. We evaluate four of our approaches ($MT_s^m, MT_s, FM_j^m, FM_j$), a \textit{separate} and a \textit{joint} adaptation strategy, with and without masking. For the $MT$ approaches we use blurring and additive noise as augmentations, and for $FM$ stronger blurring, noise and contrast shifts as strong augmentations. We compare to the following baseline methods: $UNet$, $PUNet$, and a proposal-based segmentation method from \cite{livecell} (LIVECell), which were trained on the respective source domain and applied to the target domain without any adaptation. In addition, we also use two simple domain adaptation strategies: $PUNet_{trg}$ uses the pseudo-labels predicted by the source PUNet for training a new model (with randomly initialized weights), $PUNet_{trg}^m$ is a similar approach that uses the consensus response for masking during training.

\begin{table*}[h]
\centering
\caption{Results for cell segmentation in LIVECell. Methods above the midline do not use any adaptation, methods marked in \textit{italics} were trained by us. Results show average dice score over all test images. The best method(s) for each domain are marked in \textbf{bold} and the second best method(s) in \underline{underline}.} \label{tab1}
\begin{tabular}{lrrrrrrrrr}
\toprule
Method & A172 & BT474 & BV2 & Huh7 & MCF7 & SHSY5Y & SkBr3 & SKOV3  & Average \\
\midrule
$UNet$ & 0.87 & 0.82 & 0.68 & 0.71 & 0.85 & 0.81 & 0.82 & 0.86 & 0.803 \\
$PUNet$ & 0.87 & 0.83 & 0.49 & \underline{0.79} & 0.86 & 0.79 & 0.72 & \underline{0.88}  & 0.779 \\
LIVECell & 0.84 & 0.81 & 0.30 & \underline{0.79} & 0.75 & 0.79 & 0.81 & 0.87 & 0.745 \\
\midrule
$PUNet_{trg}$ & \underline{0.88} & 0.84 & 0.51 & 0.81 & 0.86 & 0.79 & 0.73 & \underline{0.88} & 0.788 \\
$PUNet_{trg}^m$ & \textbf{0.89} & \underline{0.86} & 0.54 & 0.82 & 0.87 & 0.84 & 0.73 & \textbf{0.89} & 0.805 \\
$MT_s$ & 0.87 & 0.83 & 0.54 & \underline{0.79} & 0.87 & 0.81 & 0.70 & 0.87 & 0.785 \\
$MT_s^m$ & \underline{0.88} & \underline{0.86} & 0.76 & 0.83 & \textbf{0.89} & \textbf{0.88} & 0.69 & \underline{0.88} & 0.834 \\
$FM_j$ & \underline{0.88} & \textbf{0.88} & \underline{0.81} & \textbf{0.88} & \underline{0.88} & \underline{0.86} & \textbf{0.87} & 0.86 & \textbf{0.865} \\
$FM_j^m$ & 0.86 & \textbf{0.88} & \textbf{0.82} & \textbf{0.88} & \underline{0.88} & 0.81 & \underline{0.85} & 0.87 & \underline{0.856} \\
\bottomrule
\label{livecell_da}
\end{tabular}
\end{table*}

We see that some adaptation tasks (A172, BT474, SHSY5Y, SKOV3) are easy and all methods perform similar. For the more difficult tasks (BV2, Huh7, SkBr3) the self-training approaches, in particular $FM_j$ and $FM_j^m$, significantly improve the results. We see a clear advantage of masking for $MT$, but no significant effect on $PUNet_{trg}$ and $FM_j$. Qualitative results are shown in Fig.~\ref{fig:da-livecell}.

We demonstrate the potential of our approach for mitochondria segmentation in EM, using \textit{MitoEM} \cite{mitoem} as source dataset. It contains two large volumes with instance labels that we binarize for semantic segmentation. We use the datasets from \textit{Lucchi} \cite{lucchi}, \textit{VNC} \cite{vnc}, and \textit{UroCell} \cite{urocell} as target domains. Here, we treat all datasets as 2d segmentation tasks by considering the sections of the respective volumes as individual images. Tab.~\ref{em_da} shows the results for all our adaptation strategies, with source models $UNet$ and $PUNet$ trained on \textit{MitoEM} as baselines. Our adaptation methods significantly improve the results.
We also observe differences depending on the domain gap. For \textit{VNC} and \textit{Lucchi}, which exhibit a smaller domain gap, as seen by the relatively good result of source models $UNet$ and $PUNet$, the masking based methods $MT_s^m$, $MT_s^w$,$FM_s^m$, and $FM_s^w$ perform particularly well, with joint training counterparts showing similar performance, except for $FM_j^m$. For \textit{UroCell}, which has a much larger domain gap, $MT_j$ outperforms other approaches. Presumably the masks are unreliable in this case due to the bad prediction quality of the initial source model, and joint training prevents collapse due to bad pseudo-labels. Conversely, FixMatch approaches perform poorly overall in this setting; possibly due to the fact that strong augmentations prove detrimental for poor source models.

\begin{table}[h]
\centering
\caption{Results for mitochondrion segmentation in EM. Methods above the midline do not use any adaptation, methods marked in \textit{italics} were trained by us. Results show average dice score over all test images. The best method(s) for each domain are marked in \textbf{bold} and the second best method(s) in \underline{underline}.}
\label{em_da}
\begin{tabular}{lrrrr}
\toprule
Method & VNC & Lucchi & UroCell & Average\\
\midrule
$UNet$      & 0.67  & 0.871 & 0.239 & 0.593 \\
$PUNet$     & 0.496 & 0.809 & 0.013 & 0.439 \\
\midrule
$MT_s$      & 0.769 & 0.867 & 0.015 & 0.550 \\
$MT_s^m$    & \textbf{0.811} & 0.825 & 0.277 & \underline{0.638} \\
$MT_s^w$    & 0.789 & 0.86  & \underline{0.307} & \textbf{0.652} \\
$MT_j$      & 0.522 & 0.868 & \textbf{0.45} & 0.613 \\
$MT_j^m$    & 0.736 & 0.708 & 0.133 & 0.526 \\
$MT_j^w$    & 0.736 & \textbf{0.891} & 0.211 & 0.613 \\
$FM_s$      & 0.788 & 0.88  & 0.104 & 0.591 \\
$FM_s^m$    & \underline{0.797} & 0.883 & 0.104 & 0.595 \\
$FM_s^w$    & 0.786 & \underline{0.887} & 0.103 & 0.592 \\
$FM_j$      & 0.545 & \textbf{0.891} & 0.104 & 0.513 \\
$FM_j^m$    & 0.402 & 0.465 & 0.113 & 0.327 \\
$FM_j^w$    & 0.737 & 0.857 & 0.103 & 0.566 \\
\bottomrule
\end{tabular}
\end{table}

We also demonstrate the potential for medical imaging, using three datasets for X-Ray lung segmentation: \textit{NIH} \cite{xlsor}, \textit{Montgomery} \cite{montgomery} and \textit{JSRT} \cite{jsrt}. The latter contains two set-ups, so we obtain 4 domains and study the adaptation between all pairs. We only study $MT$ methods, and compare to the source $UNet$ and $PUNet$ baselines. Tab.~\ref{xray_da} shows the results for all our adaptation strategies. Our methods improve results for most settings, masking is beneficial for $MT_j$ and does not have a significant effect for $MT_s$. Note that \textit{JSRT2} consists of inverted X-Ray images and presents a hard adaptation task. While none of the methods succeed, $MT_j$ shows better relative performance.

\begin{table}[h]
\centering
\caption{Results for lung segmentation in X-Ray. Methods above the midline do not use any adaptation, methods marked in \textit{italics} were trained by us. Results show average dice score over all test images. The best method(s) for each domain are marked in \textbf{bold} and the second best method(s) in \underline{underline}.}
\label{xray_da}
\small
\begin{tabular}{lrrrrr}
\toprule
Method & NIH & Montgomery & JSRT1 & JSRT2 & Average \\
\midrule
$UNet$   & 0.77 & \underline{0.78} & 0.72 & 0.08 & 0.588 \\
$PUNet$  & 0.76 & 0.75 & \underline{0.73}& \underline{0.12}& 0.590 \\
\midrule
$MT_s$   & \underline{0.78} & 0.77 & \textbf{0.77} & \underline{0.12} & 0.610 \\
$MT_s^m$ & \underline{0.78} & 0.77 & \textbf{0.77} & 0.11 & 0.608 \\
$MT_j$   & \underline{0.78} & 0.77 & 0.69 & \textbf{0.21}   & \underline{0.613} \\
$MT_j^m$ & \textbf{0.82} & \textbf{0.80} & \underline{0.73} & \textbf{0.21} & \textbf{0.640} \\
\bottomrule
\end{tabular}
\end{table}

\subsection{Ablation Studies}

We study pseudo-label filtering (masking $m$, weighting $w$, no filtering) in more detail. Tab.~\ref{tab_weighting} shows the results for the difficult LIVECell adaptation settings. We see that filtering consistently improves results, except for $FM_j$ where it does not have a significant effect. In most cases masking yields better results than weighting, except for $MT_j$. In Tab.~\ref{tab_strategies} we investigate the two adaptation strategies. We see that joint training, in particular $FM_j$ yields overall better results, $FM_s$ yields inferior results and we have observed unstable training for this method.

\begin{table}
\centering
\caption{Pseudo-label filtering for cell-types in LIVECell with high domain shift. Results show average dice score over all test images. The best method(s) for each domain are marked in \textbf{bold} and the second best method(s) in \underline{underline}.}
\label{tab_weighting}
\begin{tabular}{lrrr}
\toprule
Method & BV2 & Huh7 & SkBr3 \\
\midrule
$MT_s$ & 0.54 & 0.79 & 0.70 \\
$MT_s^m$ & 0.76 & 0.83 & 0.69 \\
$MT_s^w$ & 0.65 & 0.79 & 0.74 \\
$MT_j$ & 0.64 & 0.68 & 0.59 \\
$MT_j^m$ & \textbf{0.83} & 0.75 & 0.63 \\
$MT_j^w$ & 0.78 & \underline{0.85} & 0.81 \\
$FM_j$ & 0.81 & \textbf{0.88} & \textbf{0.87} \\
$FM_j^m$ & \underline{0.82} & \textbf{0.88} & 0.85 \\
$FM_j^w$ & 0.69 & 0.77 & \underline{0.86} \\
\bottomrule
\end{tabular}
\end{table}

\begin{table}
\centering
\caption{Domain adaptation strategies for cell-types in LIVECell with high domain shift. Results show average dice score over all test images. The best method(s) for each domain are marked in \textbf{bold} and the second best method(s) in \underline{underline}.}
\label{tab_strategies}
\begin{tabular}{lrrr}
\toprule
Method & BV2 & Huh7 & SkBr3 \\
\midrule
$MT_s$ & 0.54 & 0.79 & 0.70 \\
$MT_s^m$ & 0.76 & \underline{0.83} & 0.69 \\
$MT_j$ & 0.64 & 0.68 & 0.59 \\
$MT_j^m$ & \textbf{0.83} & 0.75 & 0.63 \\
$FM_s$ & 0.43 & 0.74 & 0.74 \\
$FM_s^m$ & 0.52 & 0.79 & 0.70 \\
$FM_j$ & 0.81 & \textbf{0.88} & \textbf{0.87} \\
$FM_j^m$ & \underline{0.82} & \textbf{0.88} & \underline{0.85} \\
\bottomrule
\end{tabular}

\end{table}

\section{Discussion}

We introduce a domain adaptation framework that uses probabilistic segmentation (PUNet \cite{punet}) to generate confidence estimates, which allows us to filter the teacher pseudo-labels in the target domain and improve training for new unlabeled data.
We explore two domain adaptation regimes: i) a separate source training and subsequent target adaptation (two stage training), and ii) joint training on source and target (single stage training).
Such self-training strategies are widely used in natural images to improve domain generalization using computer vision. Our work systematically evaluates these approaches in biomedical image segmentation across three challenging scenarios with a large domain shift. 

Among the probabilistic domain adaptation methods we study, we see fluctuating performance depending on the segmentation task being studied. Nevertheless, we can identify a few trends: \textit{Mean Teacher} based approaches generally are more stable than \textit{FixMatch}, i.e. the improvements with the former are more consistent, even though the ranking of methods vary per task. Furthermore, the advantage of joint training over the two-stage approach (separate training), which is easier to use in practice as it does not require the source data for the adaptation training, is unclear.
Pseudo-label filtering via \textit{consensus masking} leads to better results in most cases, except for very large domain gaps.
Following all these observations, we recommend using mean teacher training with two stages and with confidence masking or weighting, corresponding to $MT_s^m$ or $MT_s^w$ in our nomenclature, as the best two methods for use in practice.

We see the potential to improve our methods via better probabilistic segmentation (e.g. MedSegDiff \cite{medsegdiff}) and better strong augmentations, e.g. \textit{CutMix} \cite{cutmix}, which has shown success in semi-supervision \cite{yang_revisiting_2022}.
Such probabilistic segmentation methods could also enable uncertainty-aware instance segmentation and domain adaptation, for example to learn and adapt foreground and boundary representation of cells, improving markers for seeded watershed to generate an instance segmentation.

Furthermore, our adaptation methods could be combined with recent trends to build foundation models for biomedical segmentation tasks \cite{micro-sam, cellpose-sam}, where these foundation models could function as a much stronger source model that could then be improved on target domains using domain adaptation strategies similar to the ones we have introduced here, potentially also using distillation for smaller task-specific models.

\section{Acknowledgments}
We would like to express their gratitude to Sartorius AG for support of this research through the Quantitative Cell Analytics Initiative (QuCellAI) and would also like to extend our thanks to all partners involved in the initiative for their contributions and valuable insights.
The work of Anwai Archit was funded by the Deutsche Forschungsgemeinschaft (DFG, German Research Foundation) - PA 4341/2-1.
Constantin Pape is supported by the German Research Foundation (Deutsche Forschungsgemeinschaft, DFG) under Germany’s Excellence Strategy - EXC 2067/1-390729940.
This work is supported by the Ministry of Science and Culture of Lower Saxony through funds from the program zukunft.niedersachsen of the Volkswagen Foundation for the ’CAIMed – Lower Saxony Center for Artificial Intelligence and Causal Methods in Medicine’ project (grant no. ZN4257).
We gratefully acknowledge the computing time granted by the Resource Allocation Board and provided on the supercomputer Emmy at NHR@Göttingen as part of the NHR infrastructure, under the project nim00007.

{
    \small
    \bibliographystyle{ieeenat_fullname}
    \bibliography{main}
}

\clearpage

\appendix

\section{Definition of Terms} \label{app_definitions}

We address the problem of unsupervised domain adaptation (UDA). This problem is closely related to supervised domain adaptation (SDA) and semi-supervised learning (SSL). In SSL problems both labeled and unlabeled training data is given, with the goal to learn a "better" model using all available data compared to using fully supervised learning on just the labeled data. Both labeled and unlabeled data are from the same domain (same data distribution). In the case of UDA a source domain with labels and a target domain without labels is given. Source and target domain have different data distributions, corresponding e.g. to different imaging devices or different experimental conditions in biomedical applications. The goal of UDA is to train a model that solves the same task (e.g. cell segmentation) on the target domain as on the source domain. The case of SDA is very similar, but the target domain is partially labeled (i.e. annotations are provided for a subset of the target samples).
This discussion shows that all three settings are similar, with the distinction being the data distributions for labeled and unlabeled data:
In SSL both labeled and unlabeled data come from the same data distribution and in UDA they come from two different data distributions (source and target data distribution). In SDA the labeled data comes from both source and target distribution (usually with the fraction of target data being significantly smaller) and the unlabeled data comes only from the target distribution. Consequently self-training methods can be generalized to all three learning problems, as has recently been demonstrated by \textit{AdaMatch} \cite{adamatch}. 

We make use of self-training with pseudo-labels to address UDA. These terms are sometimes used with slightly different meanings in the literature.  
Here, we use self-training to describe methods that use a version of the model being trained to generate predictions on unlabeled data, which are then used as targets in an unsupervised loss function to again train the model. This can be understood as a student-teacher set-up, with the teacher being a version of the student (e.g. through EMA of weights or weight sharing). We use the term pseudo-labeling to describe the process of transforming the teacher predictions into targets for the unsupervised loss function, e.g. by post-processing or filtering (masking or weighting) them. Note that the literature sometimes distinguish between pseudo-labeling when the likeliest prediction is used as hard target in the unsupervised loss, and consistency regularization when the softmax output (more generally output after the last activation) is used as target. See for example \url{https://lilianweng.github.io/posts/2021-12-05-semi-supervised/} for an in-depth discussion. However, this distinction is minor in practice and can be incorporated in the pseudo-labeling post-processing in our formulation (the function $p$ in Eq.~\ref{eq_pseudolabels}). Hence, we do not make this distinction throughout the paper. 

\section{Probabilistic Domain Adaptation: Training Strategies} \label{app_pseudocode}

We implement two different approaches to domain adaptation: \textit{joint} training (model is trained jointly on labeled source and unlabeled target data, using a supervised and unsupervised loss function) and \textit{separate} training (model is first pre-trained on the labeled source data, using only the supervised loss, and then fine-tuned on the unlabeled target data, using only the unsupervised loss). Here, we show pseudo-code for the two training routines, for \textit{joint} training in Alg.~\ref{alg_1} and for \textit{separate} training in Alg.~\ref{alg_2}. For simplicity we omit validation, which is performed using the target data in both cases. Here, we use the same loss function $l$ for both the supervised and unsupervised loss.

\begin{algorithm}[h]
\SetAlgoLined
\KwIn{Labeled training data \{($x_s$, $y_s$)\}, unlabeled training data \{$x_u$\}, teacher and student models $s$, $t$, number of iterations $N$}
Initialize parameters of $s$ and $t$\;
\For{$i \gets 1$ to $N$}{
Sample mini-batch ($x_s^i$, $y_s^i$) and $x_u^i$\;
Compute supervised loss $L_s = l(s(x_s^i), y_s^i)$\;
Sample augmentations $\tau_s$ and $\tau_t$\;
Compute pseudo-labels $\hat{y} = t(\tau_t(x_u))$\;
Compute unsupervised loss $L_u = l(f(s(\tau_s(x_s)), \hat{y}))$\;
Compute gradients, update parameters of $s$ based on $L_s + L_u$\;
Update parameters of $t$ from $s$;
}
\caption{Pseudo code for the joint training strategy.}
\label{alg_1}
\end{algorithm}

\begin{algorithm}[h]
\SetAlgoLined
\KwIn{Unlabeled training data \{$x_u$\}, pre-trained model $s$, number of iterations $N$}
Copy model $t$ from $s$\;
\For{$i \gets 1$ to $N$}{
Sample mini-batch $x_u^i$\;
Sample augmentations $\tau_s$ and $\tau_t$\;
Compute pseudo-labels $\hat{y} = t(\tau_t(x_u))$\;
Compute unsupervised loss $L_u = l(f(s(\tau_s(x_s)), \hat{y}))$\;
Compute gradients, update parameters of $s$ based on $L_u$\;
Update parameters of $t$ from $s$;
}
\caption{Pseudo code for the separate training strategy. (Only the adaptation stage on the target domain; source training follows regular supervised learning.)}
\label{alg_2}
\end{algorithm}

The two self-training approaches we implement, \textit{MeanTeacher} and \textit{AdaMatch} correspond to different choices for the teacher and student augmentations $\tau_s$ and $\tau_t$ as well as the teacher update scheme. For \textit{MeanTeacher} both $\tau_s$ and $\tau_t$ are sampled from a distribution of weak augmentations and the weights of $t$ are the EMA of $s$. For \textit{FixMatch} $\tau_s$ is sampled from a distribution of strong augmentations and $\tau_t$ from a distribution of strong augmentations, $s$ and $t$ share weights. The different pseudo-label filtering approaches are realized by different choices for $f$, where \textit{consensus masking} corresponds to only computing gradients for pixels that have a value of 1 in the consensus response (see Eq.~\ref{eq_consensus}), \textit{consensus weighting} corresponds to weighting the loss by the consensus response. In the case of no filtering $f$ is the identity.

\section{Implementation} \label{app_method_details}

We use the same UNet and PUNet architecture for all experiments, using an encoder-decoder architecture following the respective implementations of \cite{unet} and \cite{punet}. We increase the number of channels from 64 to 128, 256 and 512 in the encoder, and decrease it accordingly in the decoder. We use a 2d segmentation network, hence both architectures make use of 2d convolutions, 2d max-pooling and 2d upsampling operations. The UNet is trained using the Dice Error (1. - Dice Score) as loss function. For the PUNet we use a similar formulation for the loss function as in \cite{punet}, but use the Dice Error for the reconstruction term instead of the cross entropy. We use a dimension of 6 for the latent space predicted by the prior and posterior net of the PUNet. We use the Adam optimizer, relying on the default PyTorch parameter settings, except for the learning rate, and we use the \textit{ReduceLROnPlateau} learning rate scheduler. For \textit{joint} and source model trainings we train for 100k iteration, for the second stage of \textit{separate} trainings we train for 10k iterations.
We use different patch shapes, batch sizes and learning rates depending on the dataset and method; these values were determined by exploratory experiments.

For LIVECell:
\begin{itemize}
    \item $UNet$: patch shape: (256, 256); batch size: 4; learning rate: 1e-4
    \item $PUNet$: patch shape: (512, 512); batch size: 4; learning rate: 1e-5
    \item $PUNet_{trg}$, $MT_s$: patch shape: (512, 512); batch size: 2; learning rate: 1e-5
    \item $FM_s$: patch shape: (256, 256); batch size: 2; learning rate: 1e-7
    \item $FM_j$, $MT_j$: patch shape: (256, 256); batch size: 2; learning rate: 1e-5
\end{itemize}

For mitochondria segmentation in EM:
\begin{itemize}
    \item $UNet$, $PUNet$, $MT_s$, $MT_j$, $FM_j$: patch shape: (512, 512); batch size: 4; learning rate: 1e-5
    \item $FM_s$: patch shape: (512, 512); batch size: 4; learning rate: 1e-7
\end{itemize}

For lung segmentation in X-Ray:
\begin{itemize}
    \item $UNet$: patch shape: (256, 256); batch size: 2; learning rate: 1e-4
    \item $PUNet$, $MT_s$, $MT_j$: patch shape: (256, 256); batch size: 2; learning rate: 1e-5
\end{itemize}

We use gaussian blurring and additive gaussian noise (applied randomly with a probability of 0.25, and with augmentations parameters also sampled from a distribution) as weak augmentations, and gaussian blurring, additive gaussian noise and random contrast adjustments (applied randomly with a probability of 0.5, and sampling from a wider range compared to the weak augmentations) as strong augmentations.

Our implementation is based on PyTorch. We use the PUNet implementation from \url{https://github.com/stefanknegt/Probabilistic-Unet-Pytorch}. All our code is available on GitHub at \url{https://github.com/computational-cell-analytics/Probabilistic-Domain-Adaptation}. Please refer to the README for instructions on how to run and install it.

\section{Datasets} \label{app_dataset_details}

\paragraph{LIVECell Dataset}

We use the LIVECell dataset from \cite{livecell}. This dataset contains about 5000 phase contrast microscopy images with instance segmentation ground-truth and predefined train-, test-, and validation-splits. We binarize the instance segmentation ground-truth to obtain a semantic segmentation problem. The dataset contains images of 8 different cell lines: A172, BT474, BV2, Huh7, MCF7, SHSY5Y, SkBr3 and SKOV3.
These cell lines show significant difference in appearance and morphology of cells as well as spatial distribution such as cell density and cell clustering. Hence, we treat all 8 cell types as different domains, and study the adaptation from one cell line as source domain to the seven other target domains for all 8 cell lines. The columns in Tab.~\ref{tab1} show the average dice score for one source applied to the seven target domains.

\paragraph{Mitochondria EM Segmentation}

For mitochondria segmentation in EM we use the dataset of \cite{mitoem} as source dataset. This dataset contains two EM volumes, one of human neural tissue, the other of rat neural tissue, imaged with scanning EM. Each volume contains 400 images with instance annotations for training, and 100 images with instance annotations for testing. We binarize the instance segmentation ground-truth to obtain a semantic segmentation problem. We study domain adaptation with \cite{mitoem} as source for two different target datasets. The first is \textit{Lucchi} \cite{lucchi}, which contains two volumes of tissue from the murine hippocampus imaged with FIBSEM that both contain mitochondria instance annotations. We use one of the volumes for training the domain adaptation methods (either via joint or separate training), and the other for evaluation. And \textit{VNC} \cite{vnc}, which contains two volumes from the ventral nerve cord of a fruit fly, imaged with serial section transmission EM. Only one of the two volumes contains instance annotation, it is used for evaluation, the other does not, it is used for training the domain adaptation methods (which does not require labels). \textit{UroCell} \cite{urocell} contains annotated volumes of tissue from mice urinary bladders imaged with FIBSEM that contain multiple intracellular compartment annotations, including mitochondria. We use four volumes for training the domain adaptation methods (either via joint or separate training), and reserve one volume for evaluation.

\paragraph{Lung X-Ray Segmentation}

For the lung segmentation task we use four different datasets of chest radiographs, following the experiment set-up of \cite{xlsor}. The datasets are: \textit{NIH}, which contains chest X-Ray (CXR) images with various severity of lung diseases, \textit{Montgomery} \cite{montgomery}, which contains images of patients with and without tuberculosis, and \textit{JSRT} \cite{jsrt}, which contains images of patients with and without lung nodules. The \textit{JSRT} dataset is split into two subsets: \textit{JSRT1} with normal CXR images (60 images, 50 train, 10 test), and \textit{JSRT2} with inverted CXR images (247 images, 199 train, 48 test). The \textit{NIH} dataset contains 100 images (we use 80 for training and 20 for testing) and the \textit{Montgomery} dataset contains 138 images (113 are used training, 25 for testing) respectively. All datasets contain binary lung annotations; we discard additional annotations in the case of \textit{JSRT2}.
We treat each dataset as a separate domain, and perform domain adaptation for all pairs of domains.

\end{document}